# Marathi-English Code-mixed Text Generation

Dhiraj Amin[1], Sharvari Govilkar[1], Sagar Kulkarni[1], Yash Shashikant Lalit[1], Arshi Ajaz Khwaja[1], Daries Xavier[1], Sahil Girijashankar Gupta[1]

[1] Dept. of Computer Engineering, Pillai College of Engineering, Mumbai, Maharashtra, India

*Abstract*— Code-mixing, the blending of linguistic elements from distinct languages to form meaningful sentences, is common in multilingual settings, yielding hybrid languages like Hinglish and Minglish. Marathi, India's third most spoken language, often integrates English for precision and formality. Developing code-mixed language systems, like Marathi-English (Minglish), faces resource constraints. This research introduces a Marathi-English code-mixed text generation algorithm, assessed with Code Mixing Index (CMI) and Degree of Code Mixing (DCM) metrics. Across 2987 code-mixed questions, it achieved an average CMI of 0.2 and an average DCM of 7.4, indicating effective and comprehensible code-mixed sentences. These results offer potential for enhanced NLP tools, bridging linguistic gaps in multilingual societies.

*Keywords— natural language processing, code mixed data, code mixed text generation, code-mixed Marathi*

## I. Introduction

Code-mixing is the practice of blending linguistic components from two distinct languages to construct meaningful sentences, a phenomenon frequently encountered in multilingual settings, leading to the emergence of hybrid languages such as Hinglish and Minglish. In this context, the smooth integration of linguistic elements from Indian languages and English gives rise to fresh and unique expressions. Code-mixing can be employed for a variety of reasons, such as to express identity, or simply to communicate more effectively. It can also be used to add humor or creativity to a conversation. While code-mixing can be a complex phenomenon, it is a natural and often necessary part of communication in multilingual settings. Developing robust systems for code-mixed languages is challenging due to limited resources, the informal nature of these languages, and the difficulty in collecting data.

In code-mixed languages, such as Marathi, native speakers often integrate English words into their sentences. This blending of languages allows individuals to seamlessly transition between different registers of speech. Individuals find it pragmatic to integrate English terms, phrases, or sentences into their conversations, especially when discussing topics for which there might not be immediate equivalents in their native language, or when referencing global concepts that are more commonly articulated in English. For instance, when discussing technical or professional subjects, employing English terms can lend an air of expertise and professionalism to the conversation. Conversely, using the native language can evoke comfort and informality when discussing personal or culturally nuanced matters. The use of English in combination with the native language can enhance clarity and precision, as English may offer short terms or expressions that align well with certain ideas or contexts, enriching the overall quality of communication.

> **English Sentence:**
> Sachin tendulkar is an international cricketer from India who has won the 2011 world cup played in India
>
> **Marathi Sentence:**
> सचिन तेंडुलकर हा भारतातील आंतरराष्ट्रीय क्रिकेटपटू आहे ज्याने २०११ मध्ये भारतात खेळलेला विश्वचषक जिंकला आहे
>
> **Codemx Marathi-English Sentence (Devanagari script):**
> सचिन तेंडुलकर हा भारतातील *इंटरनॅशनल क्रिकेटर* आहे ज्याने २०११ मध्ये भारतात खेळलेला *वर्ल्ड कप* जिंकला आहे
>
> **Codemx Marathi-English Sentence (Latin script):**
> Sachin tendulkar haa bharatatil *international cricketer* aahe jyane 2011 madhye bharatat khedlelaa *world cup* jinkala aahe

Figure 1.1 Illustration of English sentence, Marathi sentence and codemix Marathi-English sentence

Marathi is the 3rd most spoken language in India after Hindi and Bengali.[1] Marathi is the official language of the state of Maharashtra. Marathi texts are written using Devanagari script [2] which is the same script used for Hindi language. In bilingual and multilingual communities, the phenomenon of code mixing is commonly observed, where words from two or more languages are alternated within a sentence or even entire sentences are alternated among different languages. While most existing NLP applications are designed for single languages, there is a growing availability of a vast amount of code-mixed data on the internet nowadays. Code-mixed text for Marathi-English (Minglish) can be written in either Latin script or Devanagari script as shown in Figure 1.1.

Code-mixed annotated datasets are essential for the development of code-mixed systems in a variety of domains. These datasets are priceless tools that can be generated manually by subject-matter specialists or collected from web resources like social media platforms to gather probable occurrences of code-mixed text. In order to create code-mixed text, a source text written in a regional language must be translated into the required code-mixed form as the target text. This procedure is similar to translating text.

In addition, another approach to building code-mixed datasets is to use a parallel corpus that contains two different languages. By extracting word alignments between sentences in this corpus, specific words from the native language that are likely to be replaced with English equivalents can be identified and replaced. This method helps to create datasets with code-mixed content, which can be used to train and evaluate code-mixed systems.

The main contribution of this research lies in the development of a specialized code-mixed text generation algorithm designed specifically for the generation of Marathi-English code-mixed content. This algorithm addresses the unique linguistic challenges posed by code-mixing between Marathi and English, facilitating the generation of natural and coherent code-mixed sentences. This achievement holds significant promise for enhancing the quality and efficiency of natural language processing tools and applications in multilingual settings, particularly those involving Marathi-English code-mixing, ultimately bridging communication gaps in diverse linguistic contexts.

## II. Related Work

Text that includes words and phrases from two or more different languages is known as code-mixed text. It takes knowledge of both the grammar and syntax of each language as well as the code-mixing principles to create code-mixed text, which is a difficult undertaking. Code-mixed text can be produced using a variety of approaches.

One approach to generating code-mixed text is through code-mixed text translation, which relies on a statistical machine translation model that is trained on parallel code-mixed text data. This means that the model is given a set of text pairs, where each pair consists of a source text and a target text. The source text is in one language, and the target text is in code-mixed language. The model is then trained to translate the source text into the target text. While this method is effective, it requires a large amount of training data.

Several researchers have developed code-mixed generation models that can produce Hindi-English (Hinglish) from English or vice versa by using trained neural network-based translation models.

An approach for creating code-mixed sentences based on neural networks was introduced by Gupta et al. [3]. To produce high-quality outcomes, the technique makes use of linguistic insights, trained feature representation, and transfer learning. The approach received a BLEU score of 21.55 when it was tested on a dataset of English-Hindi code-mixed text.

Ganesh Jawahar et al. [4] introduces a technique for creating Hinglish code-mixed texts using a curriculum learning strategy. They initiate the process by fine-tuning the models initially on synthetic data and subsequently on gold code-mixed data. The synthetic code-mixing approach exhibits competitiveness when compared to conventional methods like back translation under various conditions. The mT5 model achieves the best translation performance during the curriculum learning process, scoring an astounding 12.67 on the BLEU scale.

Devansh Gautam et al. [5] introduces a supervised machine translation system for English to Hinglish using the mBART model, aiming to expand resources for various applications. Leveraging mBART's pre-training, the authors propose transliterating Roman Hindi words to Devanagari script to enhance translation accuracy. Evaluations show the value of adding Hindi translations to the input, leading to a test set BLEU score of 12.22. Fine-tuning mBART with two strategies—mBART-en and mBART-hien—yields performance improvements. The study introduces BLEU-Normalized and advocates for Code-Mixing Index (CMI) to better evaluate code-mixed outputs. Error analysis and dataset examination provide insights into model limitations and data quality.

S.V. Kedar et al. [6] proposed a model that employs Hinglish as an independent language for translation into English. A curated dataset of over 10,500 Hinglish-English pairs is prepared for training and testing. The model utilizes a unique architecture comprising input layers, LSTM units, and output layers. Experimental analysis measures the model's accuracy using the BLEU score, showing a promising result of 0.4 during testing.

Zheng-Xin Yong et al. [7] investigate multilingual LLMs with zero-shot prompting for generating code-mixed data in six South East Asian languages. ChatGPT displays notable potential, producing code-mixed text successfully (68% of the time) with explicit concept definition. Both ChatGPT and InstructGPT excel in generating Singlish, achieving a 96% average success rate across prompts. However, word choice errors lead to semantic inaccuracies. Other models like BLOOMZ and Flan-T5-XXL struggle with code-mixed content generation. While highlighting LLMs' potential, the author emphasizes cautious application and native speaker involvement for annotating and editing synthetic code-mixed data.

An alternate approach is the Matrix Language Frame (MLF) theory, which focuses on the dominant (matrix) language and the inserted (embedded) language. The inserted elements align with the matrix language's grammatical structure. However, random word insertions can result in uncommon code-mixed sentences that are seldom encountered in actual communication. Applying linguistically informed strategies for word or constituent insertions using multilingual transformer models can improve the quality of code-mixed text.

Pratapa et al. [8] tackle the complexity of training language models for code-mixed (CM) languages due to limited data and multilingual intricacies. They propose a computational method using Equivalence Constraint Theory to generate valid synthetic CM data. Combining this synthetic data with real CM and monolingual data, following a specific training curriculum, significantly reduces perplexity in RNN-based language models. Importantly, they find randomly generated CM data ineffective in reducing perplexity. The study underscores the significance of linguistics-based generation, noting the absence of consensus on CM language structure. Their research involves monolingual English and Spanish data, real CM data, and synthetic CM data.

Mohd et al. [9] address the challenges of processing code-mixed languages with limited labeled and unlabeled training data by introducing the GCM tool, which generates code-mixed data using theories like Equivalence Constraint (EC) and Matrix-Language (ML), providing a resource for NLP practitioners. The development of the toolkit involves several stages: first, the alignment stage, followed by the pre-GCM stage where the input is preprocessed, and finally the GCM stage.

Vivek et al. [10] uses the Embedded-Matrix theory to synthetically create Hinglish sentences for the HinGE dataset. They introduce two rule-based systems for generating Hinglish text, where Hindi serves as the primary language (matrix language), while English is integrated as a secondary language (embedded language). These systems utilize linguistic resources such as an English-Hindi Dictionary (77,805 word pairs), cross-lingual word embeddings, GIZA++ word alignment, script transliteration, and YAKE keyword extraction. The dictionary is expanded through VecMap and GIZA++, totaling 1,52,821 word pairs. Word-aligned code-mixing, which aligns nouns and adjectives, and Phrase-aligned code-mixing, which aligns keyphrases of up to three tokens, are two methods for creating code-mixed text. Both approaches require transliterating into Roman script and replacing aligned Hindi portions with appropriate English components.

Gupta et al. [11] introduced the Multilingual and Code-mixed Visual Question Answering (MCVQA) dataset, containing 248,349 training questions and 121,512 validation questions in Hindi and code-mixed Hinglish. To generate code-mixed questions, the process involved identifying Part-of-Speech and Named Entity tags in each question. Specific Hindi words were replaced with optimal lexical translations based on their PoS and NE tags, while remaining words were substituted with Roman transliterations. In order to improve lexical translations, an iterative disambiguation approach was used in conjunction with a statistical machine translation model that had been trained using publicly accessible English-Hindi parallel corpus data.

The Multilingual and Code-mixed Visual Question Answering (MuCo-VQA) dataset, which supports five languages—Hindi, Bengali, Spanish, German, and French—and five code-mixed settings—English-Hindi, English-Bengali, English-Spanish, English-German, and English-French—was introduced by Humair Raj Khan et al. [12]. In MuCo-VQA, code-mixed questions are generated using the matrix language frame (MLF) theory, emphasizing a dominant language (matrix language) and an inserted language (embedded language) in code-mixed sentences. English questions from the VQAv1.0 dataset are translated into foreign languages via Google machine translation. By aligning English terms with their foreign language counterparts in parallel questions (en-foreign language), the method identifies English words for substitution in foreign language questions, resulting in code-mixed questions in English with elements from the foreign languages.

Another approach is to extract code-mixed data available on the internet, primarily from social media websites, by using web scraping and an efficient algorithm that identifies whether the text is in a code-mixed pattern.

Pattabhi RK Rao et al. [13] assembled a code-mixed corpus from Indian language social media text using the Twitter API. This corpus encompasses two distinct time periods and is intended for entity extraction purposes. The training data was gathered from May to June 2015, while the test data was collected between August and September 2015. The existence of concept drift within Twitter data led to the acquisition of data from these disparate time periods. This initiative extends to making the corpus accessible in three Indian languages: Hindi, Malayalam, and Tamil code-mixed form.

Nayak et al. [14] gathered tweets with Twint, focusing on code-switched Roman script for building Hinglish dataset. They built and updated a Hindi vocabulary from this data, then used a shallow subword-based LSTM to identify sentences mixing Hindi and English. This classifier was iteratively trained in a semi-supervised way, expanding a small dataset with pseudo-labels, verified manually to yield about 44,455 sentences. They fine-tuned a BERT model on this dataset for accurate language classification, important for quality code-mixing data. Sentences with 2 Hindi and 2 English words were considered code-mixed. Keeping structure, punctuation, and emoticons, shuffled data was used for training. The final dataset had 52.93 million sentences (1.04 billion tokens), with 47.79 million sentences (944 million tokens) for training and 5.13 million sentences (99 million tokens) for validation. A Devanagari HingCorpus version was made via transliteration, maintaining sentence and token balance. The code-mixing level, measured by Mixed CMI index, was 31.21.

Garg N et al. [15] have generated a sentiment Hinglish code-mixed dataset comprising approximately 3,000 tweets. These tweets were collected using Twitter APIs and targeted keywords such as "Chandrayan," "Chandrayan2," "IndiaFails," "Pulwama," and "TeleMedicine." By utilizing these keywords, tweets containing hashtags or user mentions were accumulated within the dataset. Further refinement of the dataset is implemented by eliminating punctuation and stop words.

A brand-new dataset for code-mixed Natural Language Inference (NLI) is presented by Khanuja, Simran, et al. in [16]. Premises and hypotheses in this dataset are constructed using code-mixed Hindi-English conversations from Bollywood films. The Bollywood dataset consists of Romanized versions of 18 movie scenes with both Hindi and English subtitles. The selection of scenes with more than three turns, the exclusion of monologues, and the retention of sequences with a Code Mixing Index higher than 20% are all design considerations. This dataset has 2240 hypotheses and 400 code-mixed conversation snippet premises; it does not automatically transliterate Romanized Hindi into Devanagari.

K. Singh et al. [17] dataset comprises tweets related to five events, collected using Twitter's streaming API over a 14-day span, resulting in 1,210,543 tweets. To increase the code-mixed content proportion, they employ a language identification model, yielding 98,867 tweets that fulfill specific criteria: each tweet must have a minimum of three Hindi tokens, three English tokens, at least two contiguous Hindi tokens, and at least two contiguous English tokens. After applying this filtering, the dataset is left with 98,867 tweets adhering to these conditions. The inclusion of English tweets in the filtered is attributed to the misclassification of entities, clarifying their presence in the dataset.

### III. MARATHI-ENGLISH CODE-MIXED TEXT GENERATION

Code-mixed text generation is the process of creating sentences or text that seamlessly blend two or more languages. Generating code-mixed text combines words from multiple languages, a complex task demanding an understanding of each language's grammar and code-mixing rules. Various methods exist: one employs statistical machine translation models, trained on parallel code-mixed text data, while another follows the Matrix Language Frame (MLF) theory [11]. An alternative approach is to extract code-mixed data from the internet, mainly from social media, using web scraping and pattern recognition algorithms. These approaches aim to produce syntactically accurate and semantically meaningful code-mixed sentences. The goal is to create natural-sounding code-mixed text that reflects the unique characteristics of each language.

In order to create a dataset for Marathi-English code-mixed question answering, named entity recognition, and answer type extraction, proposed Marathi-English code-mixed text-generation algorithm is shown in Figure 3.1. The Matrix Language Frame (MLF) theory, which combines two languages using a dominating language (matrix language) and an inserted language (embedded language), serves as the foundation for the proposed code-mixed text generation process. In this context, the chosen matrix language is Marathi, while the embedded language is English [12]. The challenge of having precise NLP tools for Marathi encourages the inclination toward using English, as it already possesses established NLP tools for extracting part-of-speech tags from English text.

The proposed approach requires a parallel corpus containing English and Marathi texts to facilitate the generation of code-mixed texts. The corpus can be generated by either translating existing English content into Marathi or by translating existing Marathi content into English. This strategy draws on parallel corpora to establish linguistic alignments between languages.

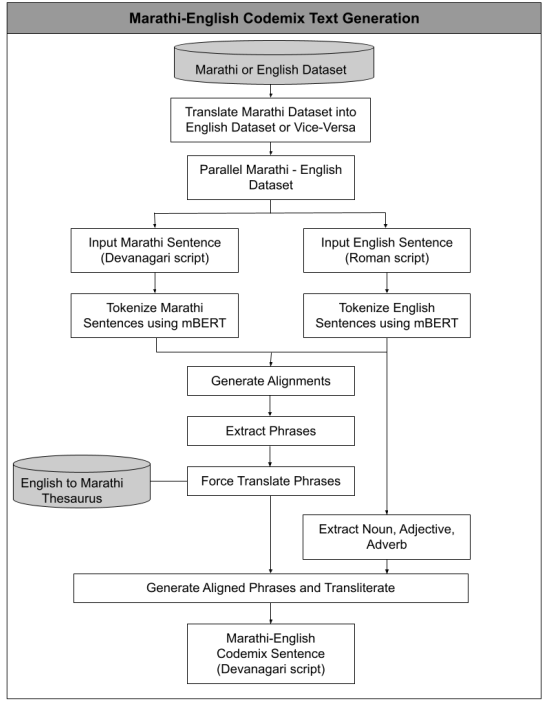

Figure 3.1 Proposed Marathi-English (Minglish) code-mixed text generation process

The algorithm processes two input sentences: one in English (Roman script) and the other in Marathi (Devnagri script). Tokens are generated for both sentences using mBERT [18], which allows for the calculation of token alignment between the two languages [18]. This process efficiently maps aligned words in English and Marathi sentences. The alignment matrix denotes the connection or mapping between words or phrases across different languages within a parallel corpus. It reveals how terms or elements in one language align with their equivalents in another language, facilitating the establishment of interconnections and associations between the two languages.

However, discrepancies in word alignment may sometimes arise, potentially affecting the precision of subsequent code-mixed text generation stages. To remedy this, a Marathi-English thesaurus is used to identify frequently substituted English words and replace them with their corresponding Marathi counterparts. As a result, these English terms assume the positions of their respective Marathi counterparts within the Marathi sentence, substantially enhancing accuracy.

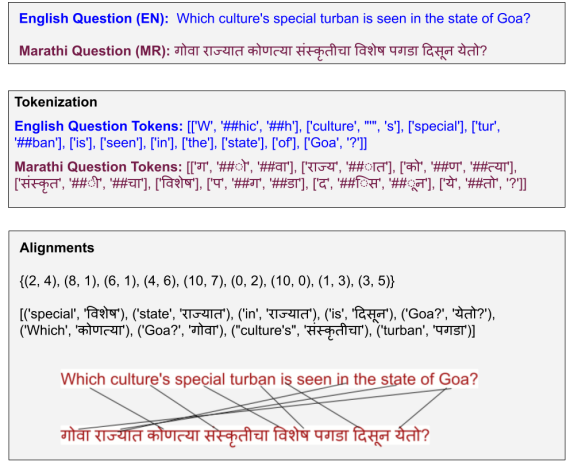

Figure 3.2 Code-Mixed text generation process (tokenization and alignment)

Phrases from the English language are selected for possible replacement in the Marathi language. Noun, adjective, and adverb phrases are carefully extracted from the English sentence using the Stanza library [19] with part-of-speech tagging. These phrases are then transliterated into the Devanagari script and seamlessly integrated with aligned Marathi words.

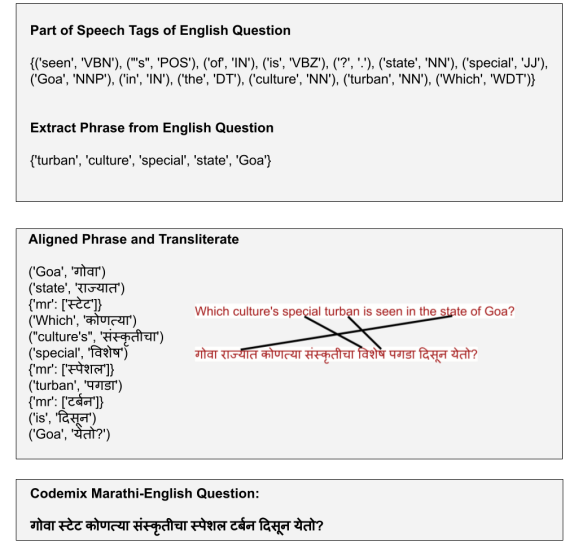

Figure 3.3 Code-Mixed text generation process (phrase extraction and transliteration)

The overall process requires the extraction of linguistic features, generation of alignment matrices,

and the careful selection of phrases to ensure the creation of coherent and meaningful code-mixed sentences. The resulting code-mixed sentences will be presented in the Devanagari script, but there is also the flexibility to generate the text in the Latin script instead of Devanagari if required. Figures 3.2 and 3.3 illustrate the process of generating code-mixed questions from parallel Marathi and English questions.

## IV. EXPERIMENT AND RESULTS

The quality of code-mixed text generation can be assessed using Code Mixing Index (CMI) and Degree of Code Mixing (DCM).

The Degree of Code Mixing (DCM) rating is a scale from 0 to 10, with 0 representing a sentence in a monolingual language with no code-mixing and 10 indicating extensive code-mixing. The evaluation is carried out by human experts who are fluent in both languages involved in code-mixing and have a high level of understanding. For a better quality code mixed sentence there should be more words of dominant language and grammatical structure should be followed of the dominant language.

The Code Mixing Index (CMI) is a metric used to quantify the amount of code-mixing in a text. It was first proposed by Gambäck and Das in 2014. [90]. The CMI is calculated as follows:

$$CMI = \frac{NM}{NT} \quad (1)$$

The CMI is determined by dividing the total number of words in the text (NT) by the number of words that are not from the dominant language (NM).

CMI has a range of values from 0 to 1. Text that is entirely code-mixed has a value of 1, while monolingual text has a value of 0.

CMI is useful for comparing code-mixing levels across various texts and monitoring changes in code-mixing within a text over time. Additionally, it allows for comparisons of code-mixing levels across different languages.

For Marathi-English code-mixed text the CMI value for a sentence will be 0 if no English words are infused in the Marathi Sentence. The value of CMI will be higher towards 1 if sufficient amounts of English words are inserted instead of Marathi words.

Marathi Sentence: भारताच्या मुख्य भूमीचा दक्षिण किनारा कोणता आहे?

English Sentence: Which is the southern coast of mainland India?

Marathi-English code-mixed sentence: भारताच्या मेनलँड भूमीचा साउथर्न कोस्ट कोणता आहे?

Number of English words in Marathi-English code-mixed sentences is 3 ( मेनलँड , साउथर्न , कोस्ट ) and there are a total of 7 words in this sentence.. The CMI for the sentence is 0.43 which indicates a quality code-mixed sentence. Similarly DCM is 10 as the sentence is readable and understandable even after code-mixing.

The evaluation process centers around a comprehensive examination of these 2987 code-mixed questions generated using the . The primary objective of this assessment is to carefully scrutinize the code-mixed nature and attributes of the questions to gain insights into their linguistic properties and effectiveness within the context of question-answering tasks.

Human evaluators are needed for evaluation using CMI and DCM. Two human evaluators, after discussing with each other, counted the number of English words inserted in each code-mixed question. This count was used to calculate the CMI and DCM values of each question. Table 5.2 presents a few sample questions from the overall dataset, along with their assigned CMI and DCM values. The blue color words are English words inserted in Marathi text.

As observed in the code-mixed question at sr. no. 2 In Table 4.1, the word "एक्सटेंट" (extent in English) does not sound natural in the sentence, thus the degree of code-mixing decreases even though the CMI is the same. In this context, the more appropriate phrase is "कुठं पर्यंत" instead of "कुठं एक्सटेंट". Although the DCM (Code-mixed Degree) value of the code-mixed question at sr. no. 2 in Table 5.2 is 10, indicating a perfect code-mixed sentence, the CMI (Code-mixed Index) value is 0.4. This implies that the CMI metric may not always provide a perfect evaluation of code-mixed sentences.

The average CMI value is 0.2 and average DCM value is 7.4 for a total 2987 code-mixed questions indicating an acceptable quality of code-mixed text for the proposed code-mixed text generation algorithm. Despite the CMI score of 0.2 suggesting minimal code-mixing (only 2 out of 10 words are code-mixed), this is considered accurate since, on average, a typical 10-word sentence may contain between 2 to 4 code-mixed words.

A DCM (Degree of Code-Mixing) value of 7.4, as determined by human assessors, reflects the level and quality of code mixing within the assessed text. While a high number of code-mixed words in a sentence can potentially diminish sentence clarity, the average DCM score of 7.4 signifies that the code-mixed questions maintain a favorable degree of quality, despite the relatively low CMI score of 0.2. This suggests that the text effectively balances code-mixing to ensure sentence comprehensibility and overall quality.

TABLE 4.1 CODE MIX INDEXING (CMI) AND DEGREE OF CODE MIXING (DCM) SCORE OF SAMPLE CODE-MIXED QUESTIONS FOR EVALUATION

| Sr. No. | Question | Number of English Tokens in CM question | Code Mix Indexing (CMI) | Degree of Code Mixing (DCM) |
|---|---|---|---|---|
| 1 | CM: पुणे शहरातील फेमस युनिव्हर्सिटी कोणते? <br><br> MR: पुणे शहरातील जगप्रसिद्ध विद्यापीठ कोणते? <br><br> EN: Which is the world famous university in Pune city? | 2 | 0.4 | 10 |
| 2 | CM: टेम्परेचर कुठं एक्सटेंट खाली येतं? <br><br> MR: तापमान कुठं पर्यंत खाली येतं? <br><br> EN: To what extent does the temperature drop? | 2 | 0.4 | 7 |
| 3 | CM: अनेक विद्यापीठांच्या करीकुलम कोणत्या अँथॉलॉजि समावेश करण्यात आला? <br><br> MR: अनेक विद्यापीठांच्या अभ्यासक्रमात कोणत्या काव्यसंग्रहाचा समावेश करण्यात आला? <br><br> EN: Which anthology was included in the curriculum of many universities? | 3 | 0.25 | 9 |

A parallel corpus of the most likely code-mixed words between English and Marathi, as well as the generation of better word alignment using transformer models, significantly contribute to the process of generating better code-mixed text.

CONCLUSION

In a multilingual landscape where code-mixing is a common practice, the research presented here addresses the need for robust systems in code-mixed languages like Marathi-English. By introducing a code-mixed text generation algorithm rooted in linguistic principles and parallel corpora, this study makes strides in enhancing the quality of code-mixed content. The evaluation results, with a CMI score of 0.2 and an average DCM score of 7.4, showcase the algorithm's effectiveness in producing natural and comprehensible code-mixed sentences. This research lays the foundation for improved NLP tools and applications catering to code-mixed languages, bridging linguistic diversity and communication gaps in multilingual societies. The promising results highlight the potential for further advancements in code-mixed text generation and applications.